\documentclass[11pt,twocolumn]{article}
\usepackage{pslatex, graphicx, amssymb, amsmath}
\usepackage{booktabs}
\usepackage{multirow}
\usepackage{colortbl}
\usepackage{amsbsy}
\usepackage{url}
\def\abstract{
\typeout{Abstract}
 {\bf Abstract} 
}

\begin{document}
\title{Generalization Boundaries of Fine-Tuned Small Language Models for Graph Structural Inference}

\author{Michal Podstawski \\ NASK National Research Institute \\ Warsaw, Poland}

\maketitle

\begin{abstract}
Small language models fine-tuned for graph property estimation have
demonstrated strong in-distribution performance, yet their
generalization capabilities beyond training conditions remain poorly
understood. In this work, we systematically investigate the boundaries of structural inference in fine-tuned small language models along two generalization axes - graph size and graph family distribution - and assess domain-learning capability on real-world graph benchmarks. Using a controlled experimental setup with
three instruction-tuned models in the 3-4B parameter class and two
graph serialization formats, we evaluate performance on graphs substantially larger than the training range and across held-out random graph families. Our results show that fine-tuned models maintain strong ordinal consistency across structurally distinct graph families and continue to rank graphs by structural properties on inputs substantially larger than those seen during training, with distinct architecture-specific degradation profiles. These findings delineate where fine-tuned small language models generalize reliably, providing empirical grounding for their use in graph-based reasoning tasks.
\end{abstract}

\section{Introduction}

Graphs are a fundamental representation for modeling complex relational
systems, and their structural properties - density, clustering,
diameter, degree distribution - are routinely computed using
well-established algorithms. For isolated graph analysis tasks, exact
computation is both sufficient and efficient. However, a growing line
of work investigates whether transformer-based language models can
reason about graph structure directly from serialized text
representations~\cite{wang2024,fatemi2024,guo2023}. This line of
inquiry is motivated not by a need to replace exact algorithms, but by
two complementary questions: whether language models develop genuine
structural understanding of graph-encoded inputs, and whether such
understanding could support approximate graph reasoning within
text-based workflows where specialized graph libraries are unavailable
or impractical, such as when graphs are embedded in natural language
descriptions, documents, or multi-modal contexts.


Recent studies have shown that small language models (SLMs), 
when fine-tuned via parameter-efficient methods, can estimate 
graph-theoretic properties directly from textual representations 
with substantial accuracy~\cite{podstawski2026tinygraphestimator}. These results establish that fine-tuned SLMs are competent graph property estimators under controlled conditions. Yet competence within the training distribution is a necessary but insufficient condition for practical utility. Before fine-tuned SLMs can be considered reliable tools for approximate graph reasoning, a prior question must be answered: \emph{where do they stop
working?}

This question has no established answer. Existing evaluations are
conducted under in-distribution conditions - matched graph families,
matched size ranges, matched domains - leaving the generalization
boundaries of fine-tuned models almost entirely uncharacterized. It
remains unknown whether structural representations learned from one
family of random graphs transfer to another or how rapidly estimation
quality degrades as graphs grow beyond the training range. Without this characterization, practitioners lack the empirical foundation needed to assess when fine-tuned SLMs can be trusted and when they cannot.

This work provides that foundation. We systematically study fine-tuned small language models for graph property estimation under complementary conditions: cross-distribution transfer between structurally distinct random graph families, size generalization beyond the training range, and learning from real-world graph domains. 

Our main findings are as follows:
\begin{itemize}
    \item Fine-tuned SLMs maintain strong ordinal consistency across
    structurally distinct random graph families, indicating that learned representations capture structural properties common across graph topologies rather than memorizing family-specific distributions.

    \item Structural reasoning degrades gracefully with increasing
    graph size. The three evaluated architectures exhibit qualitatively distinct degradation profiles, suggesting that architecture selection is a meaningful design choice for size generalization.

    \item Adjacency-list serialization consistently outperforms edge-list encoding across all experimental conditions, with the performance gap widening at larger graph sizes. Representation format emerges as a persistent factor robust to training conditions.

    \item Per-property analysis reveals a locality gradient in
    structural understanding: node-level properties are estimated
    most reliably, local-structural properties form a middle tier,
    and global combinatorial properties represent a genuine boundary of inference.
\end{itemize}


Together, these findings provide empirical grounding for the use of fine-tuned small language models in graph-based reasoning tasks and identify concrete limits on when they should and should not be relied upon.

\section{Related Work}

\subsection{Graph Reasoning in Language Models}

A growing body of work investigates the ability of language models to 
perform reasoning over graph-structured data. Early studies explored 
whether large language models (LLMs) can solve graph-related tasks 
such as shortest-path computation, connectivity analysis, and cycle 
detection when graphs are presented as text~\cite{wang2024, 
fatemi2024}. These works reveal both emergent competence and 
significant limitations, particularly for tasks requiring multi-hop 
reasoning or global structural analysis~\cite{guo2023}.



While these studies focus on zero-shot and few-shot prompting of general-purpose LLMs, a complementary line of work has shown that fine-tuned small language models can estimate a wide range of graph-theoretic properties - including density, clustering coefficients, and chromatic number - from serialized representations, substantially outperforming zero-shot baselines~\cite{podstawski2026tinygraphestimator}. The present study builds on this line of work by examining generalization boundaries of fine-tuned models across graph size, family distribution, and real-world graph domains.

\subsection{Graph Serialization and Representation}

The question of how to present graph-structured data to transformer-based language models has received sustained attention. Since transformers lack explicit structural inductive bias~\cite{vaswani2017attention}, graphs must be serialized into token sequences, and several formats have been explored in the literature, most commonly edge-list and adjacency-list encodings~\cite{guo2023}. Fatemi et al.~\cite{fatemi2024} compare multiple graph encodings in the context of zero-shot prompting of large language models, finding that the choice of representation measurably affects task performance. Analogous comparisons under fine-tuning conditions, and under distribution shift, remain comparatively less explored.

A separate line of work develops graph-native transformer architectures such as Graphormer~\cite{ying2021transformers} and GPS~\cite{rampasek2022recipe}, which introduce structural encodings and attention biases that preserve permutation invariance. These models address structural inference through architectural design rather than textual representation, and are orthogonal to the question of how general-purpose language models process serialized graph inputs.

\subsection{Generalization in Fine-Tuned Language Models}

Parameter-efficient fine-tuning methods such as LoRA
have enabled effective specialization of compact language models for 
structured reasoning tasks~\cite{hu2022lora}. 
However, the generalization properties of fine-tuned models beyond 
their training distribution remain incompletely understood. Studies 
in other domains have shown that fine-tuned models can exhibit strong 
in-distribution performance while failing to generalize to 
out-of-distribution inputs~\cite{kumar2022finetuning, wortsman2021robust}.

In the graph domain, generalization across structural distributions is particularly challenging due to the combinatorial diversity of graph topologies. GNNs guarantee permutation invariance through symmetric aggregation~\cite{xu2018powerful, kipf2017gcn}, but lack flexibility for open-ended reasoning. Language models offer greater flexibility at the cost of structural guarantees, leaving open the question of how reliably their learned representations generalize beyond the graph distributions seen during fine-tuning.



\section{Experimental Setup}

\subsection{Models}

We evaluate three small instruction-tuned transformer models: 
Llama-3.2-3B-Instruct~\cite{llama3}, Qwen2.5-3B-Instruct~\cite{qwen25}, 
and Phi-4-mini-Instruct~\cite{phi4}. These models were selected as 
representative high-performing instruction-tuned architectures in the 
3-4B parameter class, based on publicly reported rankings on the LLM 
Leaderboard~\cite{llmstats}. All models are used in their instruction-tuned variants to ensure
formatting stability during structured JSON output generation. For
brevity, the \textit{Instruct} suffix is omitted from model names
throughout the remainder of this paper.

\subsection{Fine-Tuning}
All models are fine-tuned using Low-Rank Adaptation
(LoRA)~\cite{hu2022lora} with scaling $\alpha=32$ and dropout $0.05$.
For synthetic graph experiments, LoRA rank is set to $r=32$; for
real-world graph fine-tuning, rank is reduced to $r=16$ to mitigate
overfitting on smaller dataset sizes. Training uses sequence length 2048,
batch size 2 with gradient accumulation 8, cosine learning rate schedule
with warmup ratio $0.03$, learning rate $2 \times 10^{-4}$, and 10
epochs. All experiments are conducted on NVIDIA RTX 3090 GPU.

\subsection{Graph Serialization}

Each graph is serialized in one of two formats:
\begin{itemize}
    \item \textbf{Adjacency-list (Adj):} node-wise neighbor grouping, 
    with explicit $(n, m)$ header.
    \item \textbf{Edge-list (Edge):} ordered edge pairs without 
    grouping, with explicit $(n, m)$ header.
\end{itemize}
Both formats include the number of nodes and edges in the prompt 
header. The two encodings differ only in token organization, enabling 
evaluation of locality effects independent of information content. 

\subsection{Datasets}

\subsubsection{Synthetic Graphs}
Experiments on synthetic graphs use the TinyGraphEstimator dataset~\cite{podstawski2026tinygraphestimator}, 
consisting of undirected, unweighted, connected graphs generated from three canonical random 
graph models: Erd\H{o}s--R\'{e}nyi (ER), Barab\'{a}si--Albert (BA), and Watts--Strogatz (WS). 
Node counts are uniformly sampled from $n \in [20, 30]$. The dataset comprises 1,200 training 
and 120 test graphs, evenly distributed across the three graph models. For cross-distribution 
experiments, models are trained on graphs from two families and evaluated on the held-out third, 
using 800 training graphs and 400 validation graphs per condition. For size generalization experiments, models are trained on the full 
training set and evaluated on additionally generated 
graphs of increasing size up to $n = 150$, produced using the same 
random graph models and generation methodology as the training data.

\subsubsection{Real-World Graphs}
Real-world evaluation uses a diverse selection of benchmark datasets from 
TUDataset~\cite{morris2020tudataset}: biological graphs (ENZYMES), social network 
graphs (IMDB-BINARY), and chemical graphs (NCI1); as well as molecular 
graphs from Open Graph Benchmark (OGB)~\cite{hu2020opengraph} (ogbg-moltox21). All graphs 
are converted to unweighted, undirected representations by discarding node and edge features, 
retaining only graph topology consistent with the synthetic training format. Models are 
fine-tuned separately on each dataset using an 80/10/10 train/validation/test split.

\subsection{Evaluation Metrics}

Performance is evaluated using three complementary metrics that capture
distinct aspects of estimation quality.

\textbf{Spearman rank correlation ($\rho$)} measures whether models
preserve the relative ordering of graphs with respect to each property.
For graph structural inference, ordinal consistency is a particularly
meaningful measure of understanding: a model that correctly ranks graphs
by density, clustering coefficient, or diameter demonstrates that it has
internalized the structural relationships that distinguish sparse from
dense, locally clustered from globally random, or compact from elongated
topologies. This capacity to order graphs by structural properties -
rather than merely predicting numerical values - reflects a form of
comparative structural reasoning that is robust to scale shifts and
distribution changes. Spearman $\rho$ is scale-invariant and does not
assume linearity between predicted and true values, making it directly
comparable across experimental conditions where property value
distributions differ substantially - as in the size generalization
experiments, where both the range and central tendency of properties
shift with graph size. Spearman $\rho$ serves as the primary metric
throughout this work, and is computed as:
\begin{equation}
    \rho = \frac{\sum_{i=1}^{n}(r_i - \bar{r})(s_i - \bar{s})}
               {\sqrt{\sum_{i=1}^{n}(r_i - \bar{r})^2 \sum_{i=1}^{n}(s_i - \bar{s})^2}},
\end{equation}
where $r_i$ and $s_i$ are the ranks of the $i$-th predicted and true
property values respectively, and $\bar{r}$, $\bar{s}$ are their means.

\textbf{NRMSE\textsubscript{range}} (\textbf{NR\textsubscript{rng}}) denotes the root mean squared error
normalized by the observed property range within each evaluation
condition, providing a scale-independent measure of magnitude accuracy.
Unlike Spearman $\rho$, which captures only ordinal relationships,
NRMSE\textsubscript{range} penalizes absolute prediction errors and
thus reflects whether models have learned correct value scales in
addition to correct ordering. It is computed as:
\begin{equation}
    \text{NR}_{\text{rng}} = \frac{\sqrt{\frac{1}{n}\sum_{i=1}^{n}(\hat{y}_i - y_i)^2}}{\max(y) - \min(y)},
\end{equation}
where $\hat{y}_i$ and $y_i$ are the predicted and true property values
respectively, and $\max(y) - \min(y)$ is the observed property range
within the evaluation condition.

$\mathbf{R}^{\mathbf{2}}$ reports the proportion of variance in true property
values explained by model predictions. Positive $R^2$ indicates that
the model captures meaningful variation beyond a constant mean
predictor; negative values indicate performance below this baseline,
signaling systematic prediction failure. $R^2$ is particularly
informative for cross-distribution transfer, where it distinguishes
between models that genuinely generalize magnitude estimation to
unseen graph families and those that preserve ordinal structure while
defaulting to a narrow prediction range. It is computed as:
\begin{equation}
    R^2 = 1 - \frac{\sum_{i=1}^{n}(\hat{y}_i - y_i)^2}{\sum_{i=1}^{n}(\bar{y} - y_i)^2},
\end{equation}
where $\bar{y}$ is the mean of the true property values.

The complementarity of these metrics is central to the analysis:
Spearman $\rho$ captures whether a model understands structural
differences between graphs, NRMSE\textsubscript{range} captures whether
it predicts accurate values, and $R^2$ captures whether predictions
explain meaningful variance. Conditions where Spearman $\rho$ remains
high while $R^2$ degrades - as observed in several edge-list transfer
conditions - reveal models that retain comparative structural
understanding but lose calibrated magnitude estimation, a dissociation
that informs both the practical utility and the representational
limitations of fine-tuned SLMs.

All metrics are macro-averaged across twelve evaluated graph properties:
minimum, mean, maximum, and standard deviation of node degree; graph
density; average clustering coefficient; transitivity; triangle count;
average shortest path length; diameter; chromatic number; and global
efficiency.

\section{Results}

This section presents results across three experimental conditions: 
cross-distribution transfer on synthetic graphs, size generalization, 
and real-world graph domains. We first establish in-distribution 
performance as a baseline reference.

\subsection{In-Distribution Baseline}

Table~\ref{tab:baseline} reports in-distribution performance on the
TinyGraphEstimator validation set, establishing the reference point for
all subsequent generalization experiments. All models achieve Spearman
$\rho > 0.984$ and $R^2 > 0.966$ under adjacency-list encoding,
indicating near-perfect preservation of structural ordering and strong
magnitude accuracy within the training distribution. Qwen2.5-3B
achieves the highest overall performance ($\rho = 0.987$,
$R^2 = 0.972$), though differences between models are small.
Adjacency-list encoding consistently outperforms edge-list across all
models and metrics - a pattern that persists across all subsequent
experimental conditions.


\begin{table}[ht]
\centering
\caption{Zero-shot and fine-tuned performance on the TinyGraphEstimator validation set ($n=20$--$30$). Best representation per model in \textbf{bold}.}
\label{tab:baseline}
\resizebox{\columnwidth}{!}{%
\begin{tabular}{llccccccc}
\toprule
& & \multicolumn{3}{c}{Zero-shot} & \multicolumn{3}{c}{Fine-tuned} \\
\cmidrule(lr){3-5}\cmidrule(lr){6-8}
Model & Rep. & $\rho\uparrow$ & $R^2\uparrow$ & NR$_\text{rng}\downarrow$ & $\rho\uparrow$ & $R^2\uparrow$ & NR$_\text{rng}\downarrow$ \\
\midrule
\multirow{2}{*}{Llama-3.2-3B}
 & Adj  & \textbf{0.259} & \textbf{$-$1.870} & \textbf{0.427} & \textbf{0.985} & \textbf{0.969} & \textbf{0.029} \\
 & Edge & 0.145 & $-$2.333 & 0.441 & 0.975 & 0.953 & 0.047 \\
\midrule
\multirow{2}{*}{Phi-4-mini}
 & Adj  & \textbf{0.164} & \textbf{$-$1.999} & \textbf{0.411} & \textbf{0.984} & \textbf{0.966} & \textbf{0.032} \\
 & Edge & 0.158 & $-$3.396 & 0.437 & 0.975 & 0.951 & 0.050 \\
\midrule
\multirow{2}{*}{Qwen2.5-3B}
 & Adj  & \textbf{0.261} & \textbf{$-$0.358} & \textbf{0.279} & \textbf{0.987} & \textbf{0.972} & \textbf{0.029} \\
 & Edge & 0.246 & $-$0.498 & 0.283 & 0.977 & 0.960 & 0.043 \\
\bottomrule
\end{tabular}%
}
\end{table}

\subsection{Cross-Distribution Transfer}

We evaluate cross-distribution generalization by training models on two
of the three synthetic graph families and evaluating on the held-out
third. This design tests whether fine-tuned models learn structural
representations that generalize beyond the specific topology of their
training graphs. Results are presented in Table~\ref{tab:cross_dist}.

\begin{table*}[ht]
\centering
\caption{Cross-distribution transfer results. Models trained on two graph families, evaluated on the held-out third. ER: Erd\H{o}s--R\'{e}nyi, BA: Barab\'{a}si--Albert, WS: Watts--Strogatz.}
\label{tab:cross_dist}
\resizebox{\textwidth}{!}{%
\newcolumntype{C}{>{\centering\arraybackslash}p{2.5em}}
\begin{tabular}{ll*{9}{C}}
\toprule
& & \multicolumn{3}{c}{BA+WS $\rightarrow$ ER} & \multicolumn{3}{c}{ER+BA $\rightarrow$ WS} & \multicolumn{3}{c}{ER+WS $\rightarrow$ BA} \\
\cmidrule(lr){3-5}\cmidrule(lr){6-8}\cmidrule(lr){9-11}
Model & Rep. & $\rho\uparrow$ & $R^2\uparrow$ & NR$_\text{rng}\downarrow$ & $\rho\uparrow$ & $R^2\uparrow$ & NR$_\text{rng}\downarrow$ & $\rho\uparrow$ & $R^2\uparrow$ & NR$_\text{rng}\downarrow$ \\
\midrule
\multirow{2}{*}{Llama-3.2-3B} & Adj & \textbf{0.888} & \textbf{0.647} & \textbf{0.145} & \textbf{0.946} & \textbf{0.556} & \textbf{0.158} & \textbf{0.943} & \textbf{0.697} & \textbf{0.111} \\
 & Edge & 0.861 & 0.411 & 0.184 & 0.930 & 0.291 & 0.214 & 0.882 & 0.262 & 0.175 \\
\midrule
\multirow{2}{*}{Phi-4-mini} & Adj & \textbf{0.913} & \textbf{0.733} & \textbf{0.124} & \textbf{0.952} & \textbf{0.611} & \textbf{0.139} & \textbf{0.956} & \textbf{0.798} & \textbf{0.101} \\
 & Edge & 0.863 & 0.563 & 0.153 & 0.927 & 0.231 & 0.201 & 0.907 & 0.263 & 0.180 \\
\midrule
\multirow{2}{*}{Qwen2.5-3B} & Adj & \textbf{0.901} & \textbf{0.677} & \textbf{0.133} & \textbf{0.950} & \textbf{0.469} & \textbf{0.157} & \textbf{0.950} & \textbf{0.702} & \textbf{0.116} \\
 & Edge & 0.864 & 0.534 & 0.156 & 0.917 & 0.254 & 0.455 & 0.900 & 0.063 & 0.199 \\
\bottomrule
\end{tabular}%
}
\end{table*}

Under adjacency-list serialization, all models achieve strong transfer
across all three held-out conditions. Spearman rank correlations range
from $0.888$ to $0.956$, indicating robust preservation of structural
ordering on unseen graph families. $R^2$ values are consistently
positive, ranging from $0.469$ to $0.798$, confirming that learned
representations explain a meaningful proportion of variance on held-out
distributions rather than defaulting to trivial constant prediction.

The three transfer directions reveal a consistent asymmetry across all
models. Transfer to Barab\'{a}si--Albert graphs (ER+WS$\rightarrow$BA)
and to Watts--Strogatz graphs (ER+BA$\rightarrow$WS) yields the
strongest Spearman correlations, while transfer to Erd\H{o}s--R\'{e}nyi
graphs (BA+WS$\rightarrow$ER) shows slightly reduced but still strong
performance. This pattern may reflect the higher structural regularity
of WS and BA graphs - Watts--Strogatz graphs exhibit characteristic
clustering and path length distributions, while Barab\'{a}si--Albert
graphs follow predictable degree distributions - compared to the less
constrained topology of Erd\H{o}s--R\'{e}nyi random graphs.

Among models, Phi-4-mini achieves the strongest cross-distribution
transfer overall, with Spearman correlations of $0.913$, $0.952$, and
$0.956$ across the three directions under adjacency-list encoding, and
the highest $R^2$ values in all conditions ($0.733$, $0.611$, $0.798$).
Qwen2.5-3B and Llama-3.2-3B show comparable Spearman performance but
with lower $R^2$, particularly in the ER+BA$\rightarrow$WS condition,
where Qwen2.5-3B achieves $R^2 = 0.469$ - still positive but
indicating greater difficulty with magnitude estimation on
Watts--Strogatz graphs despite strong ordinal consistency
($\rho = 0.950$).

Edge-list serialization shows substantially weaker and less consistent
transfer. 
Under ER+BA$\rightarrow$WS, all three models show severely reduced $R^2$ values under edge-list encoding (0.231--0.291), and Qwen2.5-3B additionally exhibits a substantially elevated NRMSE$_{\text{range}}$ of $0.455$, more than twice the corresponding adjacency-list value.
The
ER+WS$\rightarrow$BA condition further highlights the representation
gap: adjacency-list $R^2$ values range from $0.697$ to $0.798$, while
edge-list $R^2$ drops to $0.063$--$0.263$. This systematic contrast
is consistent with the hypothesis that adjacency-list encodings, by
grouping neighbor information per node, better support local structural
aggregation within transformer attention mechanisms - an advantage
that becomes critical when models must generalize to structurally
unfamiliar graph families.

The dissociation between Spearman and $R^2$ under edge-list encoding
is particularly informative: models can preserve approximate structural
ordering even when magnitude estimation collapses, suggesting that
ordinal and cardinal aspects of graph property inference rely on
partially distinct learned representations.

\subsection{Size Generalization}

\begin{figure*}[ht]
\centering
\includegraphics[width=\textwidth]{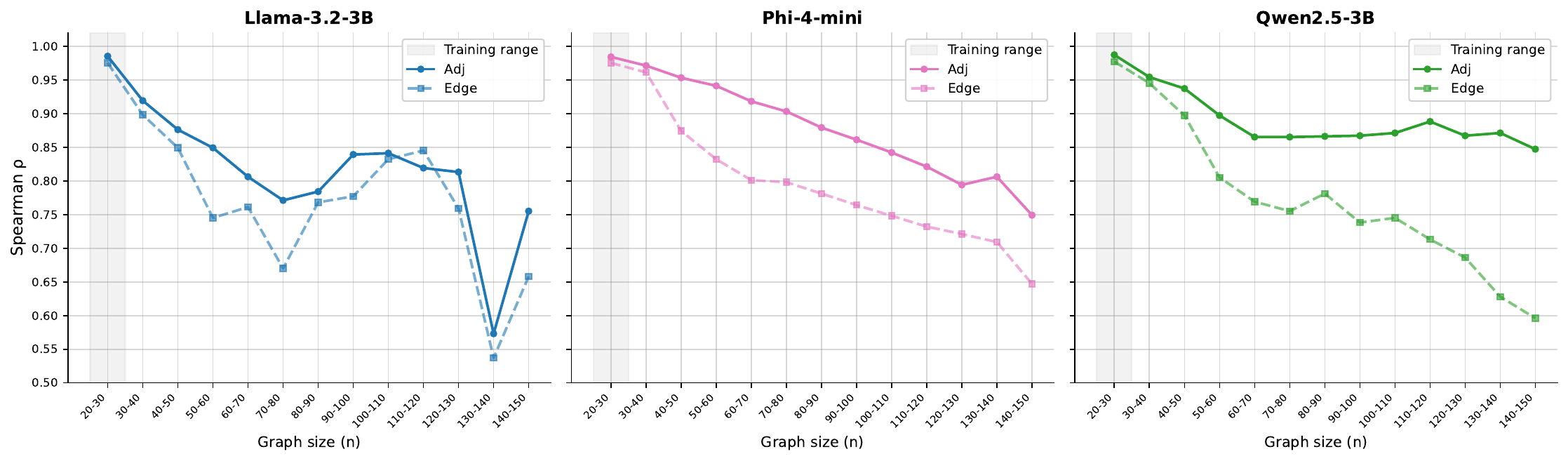}
\caption{Size generalization: Spearman~$\rho$ across increasing graph size bins for each model. Solid lines denote adjacency-list serialization, dashed lines denote edge-list. The shaded region marks the training range ($n=20$--$30$). All models show graceful degradation under adjacency-list encoding, with performance gaps between representations widening at larger graph sizes.}
\label{fig:size_gen}
\end{figure*}

We evaluate size generalization by testing models trained on $n=20$--$30$
graphs against increasingly larger graphs up to $n=150$. We report
Spearman rank correlation as the primary metric for this analysis, as
magnitude-based metrics are not directly comparable across size bins due
to changing property value distributions and normalization denominators.
Results are presented in Table~\ref{tab:size_gen} and visualized in
Figure~\ref{fig:size_gen}.

\begin{table*}[ht]
\centering
\caption{Size generalization results: Spearman~$\rho$ across graph size bins. Models fine-tuned on $n=20$--$30$ (training range). Higher is better~($\uparrow$).}
\label{tab:size_gen}
\resizebox{\textwidth}{!}{%
\newcolumntype{C}{>{\centering\arraybackslash}p{3.5em}}
\begin{tabular}{ll*{13}{C}}
\toprule
& & Train & \multicolumn{12}{c}{Out-of-distribution ($\rho\uparrow$)} \\
\cmidrule(lr){3-3}\cmidrule(lr){4-15}
Model & Rep. & 20--30 & 30--40 & 40--50 & 50--60 & 60--70 & 70--80 & 80--90 & 90--100 & 100--110 & 110--120 & 120--130 & 130--140 & 140--150 \\
\midrule
\multirow{2}{*}{Llama-3.2-3B}
 & Adj  & 0.985 & 0.919 & 0.876 & 0.849 & 0.806 & 0.771 & 0.784 & 0.839 & 0.841 & 0.819 & 0.813 & 0.573 & 0.755 \\
 & Edge & 0.975 & 0.898 & 0.849 & 0.745 & 0.761 & 0.670 & 0.768 & 0.777 & 0.832 & 0.845 & 0.759 & 0.537 & 0.658 \\
\midrule
\multirow{2}{*}{Phi-4-mini}
 & Adj  & 0.984 & 0.971 & 0.953 & 0.941 & 0.918 & 0.903 & 0.879 & 0.861 & 0.842 & 0.821 & 0.794 & 0.806 & 0.749 \\
 & Edge & 0.975 & 0.961 & 0.874 & 0.832 & 0.801 & 0.798 & 0.781 & 0.764 & 0.748 & 0.732 & 0.721 & 0.709 & 0.647 \\
\midrule
\multirow{2}{*}{Qwen2.5-3B}
 & Adj  & 0.987 & 0.954 & 0.937 & 0.897 & 0.865 & 0.865 & 0.866 & 0.867 & 0.871 & 0.888 & 0.867 & 0.871 & 0.847 \\
 & Edge & 0.977 & 0.945 & 0.897 & 0.805 & 0.769 & 0.755 & 0.781 & 0.738 & 0.745 & 0.713 & 0.686 & 0.628 & 0.596 \\
\bottomrule
\end{tabular}%
}
\end{table*}

All models maintain meaningful structural ordering well beyond the training size range under adjacency-list serialization. Phi-4-mini and Qwen2.5-3B remain above $\rho = 0.74$ across the entire evaluated range (up to $n = 150$), while Llama-3.2-3B shows more variable behavior, with performance above $\rho = 0.77$ in most size bins but a localized dip at $n = 130$--$140$. The
three models exhibit qualitatively distinct degradation profiles.

Phi-4-mini displays the most gradual and consistent degradation under
adjacency-list encoding, with Spearman $\rho$ declining nearly
monotonically from $0.984$ at the training range to $0.749$ at
$n=140$--$150$. This smooth trajectory suggests stable extrapolation of
learned structural representations, with no abrupt failure points across
the evaluated size range.

Qwen2.5-3B exhibits a distinctive plateau pattern under adjacency-list
encoding: after an initial decline from $0.987$ to $0.865$ at
$n=60$--$70$, performance stabilizes and remains between $0.847$ and
$0.888$ throughout the remainder of the evaluated range. This plateau
behavior suggests that the model's learned structural representations
transfer robustly once a moderate size gap is crossed, without further
degradation at larger scales.

Llama-3.2-3B shows more variable behavior, with a notable dip to
$\rho = 0.573$ at $n=130$--$140$ followed by partial recovery to
$0.755$ at $n=140$--$150$. This non-monotonic pattern suggests greater
sensitivity to specific graph size regimes, though performance remains
above $0.77$ for most of the evaluated range.

Edge-list serialization consistently underperforms adjacency-list across
all models and size bins, with the performance gap widening at larger
graph sizes. Under edge-list encoding, Qwen2.5-3B degrades from
$0.977$ to $0.596$, Phi-4-mini from $0.975$ to $0.647$, and
Llama-3.2-3B from $0.975$ to $0.658$ at the largest evaluated size.
This widening gap suggests that the locality advantage of adjacency-list
encoding becomes increasingly important as graphs grow beyond the
training distribution, consistent with the hypothesis that node-grouped
representations better support structural aggregation within
transformer attention mechanisms.

\begin{table*}[ht]
\centering
\caption{Real-world graph results. Models fine-tuned separately per dataset with LoRA $r=16$. Best representation per model and dataset in \textbf{bold}.}
\label{tab:realworld}
\resizebox{\textwidth}{!}{%
\newcolumntype{C}{>{\centering\arraybackslash}p{2.6em}}
\begin{tabular}{ll*{3}{C}*{3}{C}*{3}{C}*{3}{C}}
\toprule
& & \multicolumn{3}{c}{ENZYMES} & \multicolumn{3}{c}{IMDB-BINARY} & \multicolumn{3}{c}{NCI1} & \multicolumn{3}{c}{ogbg-moltox21} \\
\cmidrule(lr){3-5}\cmidrule(lr){6-8}\cmidrule(lr){9-11}\cmidrule(lr){12-14}
Model & Rep. & $\rho\uparrow$ & $R^2\uparrow$ & NR$_\text{rng}\downarrow$ & $\rho\uparrow$ & $R^2\uparrow$ & NR$_\text{rng}\downarrow$ & $\rho\uparrow$ & $R^2\uparrow$ & NR$_\text{rng}\downarrow$ & $\rho\uparrow$ & $R^2\uparrow$ & NR$_\text{rng}\downarrow$ \\
\midrule
\multirow{2}{*}{Llama-3.2-3B}
 & Adj  & \textbf{0.894} & \textbf{0.720} & \textbf{0.087} & \textbf{0.917} & \textbf{0.543} & \textbf{0.194} & \textbf{0.770} & \textbf{0.678} & \textbf{0.058} & \textbf{0.948} & \textbf{0.931} & \textbf{0.022} \\
 & Edge & 0.733 & 0.433 & 0.138 & 0.910 & 0.519 & 0.198 & 0.723 & 0.616 & 0.069 & 0.863 & 0.818 & 0.037 \\
\midrule
\multirow{2}{*}{Phi-4-mini}
 & Adj  & \textbf{0.839} & \textbf{0.582} & \textbf{0.109} & \textbf{0.913} & \textbf{0.683} & \textbf{0.120} & \textbf{0.685} & \textbf{0.480} & \textbf{0.078} & \textbf{0.859} & \textbf{0.604} & \textbf{0.055} \\
 & Edge & 0.727 & 0.481 & 0.131 & 0.868 & 0.565 & 0.146 & 0.597 & 0.443 & 0.094 & 0.781 & 0.553 & 0.116 \\
\midrule
\multirow{2}{*}{Qwen2.5-3B}
 & Adj  & \textbf{0.793} & \textbf{0.634} & \textbf{0.104} & \textbf{0.940} & \textbf{0.824} & \textbf{0.084} & \textbf{0.764} & \textbf{0.589} & \textbf{0.057} & \textbf{0.923} & \textbf{0.852} & \textbf{0.040} \\
 & Edge & 0.740 & 0.470 & 0.129 & 0.912 & 0.821 & 0.097 & 0.728 & 0.545 & 0.093 & 0.897 & 0.731 & 0.048 \\
\bottomrule
\end{tabular}%
}
\end{table*}


\begin{table*}[ht]
\centering
\caption{Per-property results on ENZYMES under adjacency-list encoding. Chromatic number estimation fails across all models (see text); a dash (—) indicates Spearman $\rho$ is undefined due to constant model predictions. Properties are grouped by locality: node-level, local-structural, and global-structural.}
\label{tab:perprop_enzymes}
\resizebox{\textwidth}{!}{%
\newcolumntype{C}{>{\centering\arraybackslash}p{2.8em}}
\begin{tabular}{l*{3}{C}*{3}{C}*{3}{C}}
\toprule
& \multicolumn{3}{c}{Llama-3.2-3B} & \multicolumn{3}{c}{Phi-4-mini} & \multicolumn{3}{c}{Qwen2.5-3B} \\
\cmidrule(lr){2-4}\cmidrule(lr){5-7}\cmidrule(lr){8-10}
Property & $\rho\uparrow$ & $R^2\uparrow$ & NR$_\text{rng}\downarrow$ & $\rho\uparrow$ & $R^2\uparrow$ & NR$_\text{rng}\downarrow$ & $\rho\uparrow$ & $R^2\uparrow$ & NR$_\text{rng}\downarrow$ \\
\midrule
\multicolumn{10}{l}{\textit{Node-level properties}} \\
Degree min           & 0.977 & 0.939 & 0.061 & 0.971 & 0.939 & 0.061 & 0.668 & 0.121 & 0.232 \\
Degree mean          & 0.978 & 0.893 & 0.067 & 0.998 & 0.884 & 0.070 & 0.821 & 0.663 & 0.118 \\
Degree max           & 1.000 & 1.000 & 0.000 & 0.964 & 0.938 & 0.056 & 1.000 & 1.000 & 0.000 \\
Degree std           & 0.852 & 0.662 & 0.134 & 0.855 & 0.681 & 0.131 & 0.651 & 0.480 & 0.167 \\
\midrule
\multicolumn{10}{l}{\textit{Local-structural properties}} \\
Density              & 0.978 & 0.865 & 0.050 & 0.871 & 0.780 & 0.063 & 0.989 & 0.938 & 0.033 \\
Triangle count       & 0.806 & 0.643 & 0.140 & 0.794 & 0.429 & 0.177 & 0.812 & 0.655 & 0.137 \\
Avg.\ clustering     & 0.792 & 0.711 & 0.124 & 0.803 & 0.522 & 0.159 & 0.835 & 0.777 & 0.108 \\
Transitivity         & 0.812 & 0.666 & 0.121 & 0.804 & 0.498 & 0.148 & 0.837 & 0.747 & 0.105 \\
\midrule
\multicolumn{10}{l}{\textit{Global-structural properties}} \\
Avg.\ shortest path  & 0.779 & 0.400 & 0.129 & 0.825 & 0.112 & 0.156 & 0.819 & 0.425 & 0.126 \\
Diameter             & 0.767 & 0.459 & 0.129 & 0.817 & $-$0.134 & 0.186 & 0.786 & 0.324 & 0.144 \\
Chromatic number     & —   & $-$0.007 & 0.206 & 0.187 & $-$0.093 & 0.224 & 0.006 & $-$0.093 & 0.224 \\
Global efficiency    & 0.879 & 0.848 & 0.058 & 0.855 & 0.593 & 0.094 & 0.879 & 0.837 & 0.060 \\
\midrule
Macro-average        & 0.894 & 0.720 & 0.087 & 0.839 & 0.582 & 0.109 & 0.793 & 0.634 & 0.104 \\
\bottomrule
\end{tabular}%
}
\end{table*}

\subsection{Real-World Domain Learning}

We evaluate whether fine-tuned SLMs can learn structural inference directly from real-world graph data by fine-tuning models separately on benchmark datasets. This experiment also assesses whether the representation findings from synthetic training extend to real-world domains. Results are presented in Table~\ref{tab:realworld}.

Fine-tuning on real-world graphs yields strong in-distribution
performance, though with greater cross-dataset and cross-model
variability than observed in synthetic experiments. Under adjacency-list encoding, Spearman correlations range from $0.685$ (Phi-4-mini on NCI1) to $0.948$ (Llama-3.2-3B on ogbg-moltox21), with $R^2$ values between $0.480$ and $0.931$. The wider performance spread compared to
synthetic in-distribution results (Table~\ref{tab:baseline}) reflects
the structural diversity of real-world graph domains.

The four datasets reveal distinct patterns across structural domains. On ENZYMES, a biological graph benchmark, the adjacency-list advantage is pronounced:
Llama-3.2-3B achieves $\rho = 0.894$ under adjacency-list versus
$0.733$ under edge-list, a gap of $0.161$ - substantially larger than
any representation gap observed in synthetic experiments. Phi-4-mini and
Qwen2.5-3B show similar patterns, with adjacency-list margins of
$0.112$ and $0.053$ respectively. The strong Adj--Edge contrast on
ENZYMES is consistent with the irregular topology of biological graphs,
where node-grouped representations may provide greater benefit for
structural aggregation. Llama-3.2-3B achieves the strongest overall
ENZYMES performance ($\rho = 0.894$, $R^2 = 0.720$), reversing the
model ranking observed in synthetic cross-distribution transfer where
Phi-4-mini led - suggesting that model-specific advantages may depend
on the structural characteristics of the target domain.

On IMDB-BINARY, a social network benchmark with larger and more
structurally uniform graphs, all models achieve higher Spearman
correlations ($0.868$--$0.940$) but the adjacency-list advantage is
narrower. Qwen2.5-3B achieves the strongest performance overall
($\rho = 0.940$, $R^2 = 0.824$), with only a modest gap between
representations ($\rho = 0.940$ vs $0.912$). This reduced
representation gap may reflect the more regular degree distributions and
community structure of social network graphs, where edge-list encoding
loses less structural information relative to adjacency-list.

The remaining two datasets reveal the full performance range across real-world domains.
On ogbg-moltox21, models achieve the strongest results across all four benchmarks: 
Llama-3.2-3B reaches $\rho = 0.948$ and $R^2 = 0.931$ under adjacency-list encoding, 
and Qwen2.5-3B achieves $\rho = 0.923$ and $R^2 = 0.852$. This strong performance 
likely reflects the structural regularity of molecular graphs, where valency constraints 
and bond topology produce consistent local degree patterns that align well with the 
properties models learn to estimate from synthetic training data. NCI1, by contrast, 
yields the weakest real-world results overall, with Spearman correlations ranging from 
$\rho = 0.685$ to $\rho = 0.770$ under adjacency-list encoding and $R^2$ values as low 
as $0.480$ for Phi-4-mini. NCI1 chemical compound graphs exhibit greater structural 
diversity and irregular topologies compared to molecular graphs, plausibly making local 
structural cues less predictive of global properties. The adjacency-list advantage remains present on both datasets, though less pronounced than on ENZYMES, consistent with the pattern that representation format provides greater benefit when target graphs have less regular structure.

These results demonstrate that small language models can effectively
learn structural inference from real-world graph data spanning diverse
domains when provided domain-appropriate training. The adjacency-list
advantage persists but varies in magnitude across domains, suggesting
that the benefit of node-grouped representations depends on the
structural regularity of the target graph family.

\subsection{Per-Property Analysis}

Table~\ref{tab:perprop_enzymes} disaggregates performance on ENZYMES by individual graph property under adjacency-list encoding, serving as a representative example of the locality gradient in structural understanding observed across datasets.

Node-level degree statistics are estimated most reliably: degree max
achieves perfect Spearman correlation for both Llama-3.2-3B and
Qwen2.5-3B, and degree mean exceeds $\rho = 0.82$ across all models.
These properties are directly readable from local neighborhoods in the
adjacency-list representation, requiring minimal cross-node integration.

Local-structural properties - density, clustering coefficient,
transitivity, and triangle count - form a middle tier, with Spearman
correlations between $0.79$ and $0.99$. These properties require
aggregation over local neighborhoods but remain grounded in pairwise
and triplet-level relationships that adjacency-list encoding makes
relatively accessible.

Global-structural properties - average shortest path length and
diameter - show the weakest performance among estimable properties,
with Spearman correlations between $0.77$ and $0.82$. These properties
require integration of path information across the entire graph, which
no serialization format makes directly available. The modest but
meaningful performance suggests that fine-tuned models can partially
infer global structure from local cues, likely through correlations
between local connectivity patterns and global path properties.


Chromatic number represents a clear failure case, with Spearman 
correlations near zero or undefined across all models. The 
consistent failure is consistent with its NP-hardness and reliance 
on global constraint satisfaction rather than local structural 
patterns, though we cannot rule out contributions from other 
factors such as the small integer range of valid colorings or the 
training label distribution. Regardless of the precise mechanism, 
the result suggests a genuine boundary: fine-tuned SLMs can learn 
properties that are either locally computable or statistically 
correlated with local structure, but struggle with properties 
requiring combinatorial global reasoning.

Global efficiency - despite being a global property defined over all
shortest paths - is estimated with relatively high accuracy
($\rho > 0.85$), likely because it correlates strongly with local
density and degree distribution, which the models estimate well.

\section{Discussion}

The results reveal a consistent and coherent picture of generalization
in fine-tuned small language models for graph property estimation.
Across synthetic cross-distribution transfer, size generalization, and
real-world graph domains, three patterns emerge with particular
regularity: fine-tuned models maintain meaningful structural ordering
well beyond their training conditions, adjacency-list serialization
consistently enables stronger performance than edge-list encoding, and
ordinal consistency degrades more gracefully than magnitude accuracy
across all out-of-distribution conditions.

\paragraph{Cross-distribution transfer and structural abstraction.}
The cross-distribution transfer results are perhaps the most surprising
finding. Despite being trained exclusively on two of three random graph
families, models achieve Spearman correlations above $0.88$ on the
held-out family under adjacency-list encoding - suggesting that
fine-tuned SLMs learn structural representations that capture properties
common across graph families, rather than memorizing family-specific
value distributions. This interpretation is supported by the
consistently positive $R^2$ values under adjacency-list encoding, which
indicate genuine variance explanation rather than trivial constant
prediction.

The transfer direction asymmetry - stronger transfer to
Watts--Strogatz and Barab\'{a}si--Albert graphs than to
Erd\H{o}s--R\'{e}nyi - is structurally interpretable. WS and BA
graphs are generated by constrained processes that produce characteristic
structural signatures: regular lattice rewiring in WS and preferential
attachment in BA. These regularities may be easier to generalize to
precisely because they are more predictable from local structural
patterns. Erd\H{o}s--R\'{e}nyi graphs, generated by independent edge
sampling, lack such local-to-global regularity, making them harder
targets for transfer from structurally constrained training
distributions.

\paragraph{Size generalization and model-specific degradation.}
The size generalization results reveal that fine-tuned SLMs can
extrapolate structural reasoning well beyond their training range, but
the character of this extrapolation differs meaningfully across
architectures. Phi-4-mini exhibits smooth, nearly monotonic degradation
- consistent with stable learned representations that degrade
gradually as input length increases. Qwen2.5-3B shows a distinctive
plateau effect, maintaining performance after an initial decline,
suggesting that its representations capture scale-invariant structural
patterns that transfer robustly once a moderate size gap is crossed.
Llama-3.2-3B displays more volatile behavior, with performance fluctuating across size bins rather than following a clean monotonic or plateau trajectory - suggesting that its learned representations are less stable under length extrapolation than those of the other two models.

These distinct degradation profiles suggest that different architectures
internalize graph structure through qualitatively different
representations. Whether these differences stem from tokenization
schemes, attention head specialization, or pretraining data composition
remains an open question, but the practical implication is clear:
architecture selection matters for size generalization, and evaluation
at a single out-of-distribution scale may not capture the full
degradation trajectory.

\paragraph{Ordinal versus cardinal estimation.}
A recurring theme across all experimental conditions is the dissociation
between ordinal and cardinal performance. In cross-distribution transfer, edge-list models maintain moderate Spearman correlations while $R^2$ drops sharply, in some cases close to zero. In size generalization, all models
preserve structural ordering far better than they preserve magnitude
accuracy. This pattern suggests that ordinal and cardinal aspects of
graph property estimation rely on partially distinct learned
representations - with ordinal consistency being more robust to
distribution shift.

This dissociation has practical implications. For applications requiring structural comparison or ranking - identifying the densest, most clustered, or most connected graph in a set - fine-tuned SLMs offer reliable out-of-distribution performance. When precise numerical predictions are required, reliable performance is confined to a narrower range, and domain-specific calibration may be necessary.

\paragraph{Representation format as a persistent factor.}
The adjacency-list advantage is the most consistent finding across all
conditions. It holds in-distribution (Table~\ref{tab:baseline}),
across graph families (Table~\ref{tab:cross_dist}), across graph sizes
(Table~\ref{tab:size_gen}), and on real-world benchmarks. The consistency across fundamentally different experimental conditions supports the interpretation that node-grouped representations provide a genuine structural benefit for transformer attention mechanisms, rather than being an artifact of a specific evaluation condition.

The widening Adj--Edge gap at larger graph sizes
(Figure~\ref{fig:size_gen}) provides additional mechanistic insight.
As graphs grow, edge-list representations distribute relational
information across an increasingly long token sequence, requiring
attention mechanisms to integrate information across wider spans.
Adjacency-list encoding mitigates this by clustering all neighbor
information for each node within a contiguous token span, effectively
reducing the attention distance required for local structural
aggregation. This locality advantage plausibly compounds with graph
size, explaining the observed divergence.

\section{Limitations and Future Work}

Several limitations constrain the scope of the conclusions presented in
this work.

First, size generalization results rely exclusively on Spearman rank
correlation, as magnitude-based metrics are not reliably comparable
across graph size bins due to changing property value distributions and
normalization denominators. While Spearman provides a robust measure of
structural ordering preservation, it does not capture absolute
prediction accuracy at larger scales. The dissociation between ordinal
consistency and magnitude accuracy observed in cross-distribution
experiments suggests that size generalization of absolute predictions
may degrade more rapidly than the Spearman results indicate.

Second, all experiments are restricted to undirected, unweighted, connected graphs, consistent with the training corpus. Directed graphs, weighted graphs, and graphs with multiple connected components introduce additional structural dimensions - edge directionality, weight distributions, component-level statistics - that are absent from the current evaluation. The generalization patterns observed here, including the adjacency-list advantage and the locality gradient in structural understanding, are likely to carry over to these broader graph classes, since they reflect properties of how transformer attention aggregates local structural information rather than properties specific to undirected connected graphs. This expectation, however, requires empirical verification.

Finally, all experiments use models in the 3--4B parameter range. Whether
the observed generalization patterns - including the distinct
degradation profiles of different architectures and the persistent
adjacency-list advantage - hold for larger or smaller architectures
remains an open question.

\section{Conclusion}

This work presents a systematic empirical study of generalization in
fine-tuned small language models for graph property estimation. We evaluate three instruction-tuned models under complementary conditions: cross-distribution transfer between synthetic random graph families, size generalization beyond the training range, and learning of structural inference from real-world graph domains.


Our results demonstrate that fine-tuned SLMs generalize meaningfully
within the synthetic domain - maintaining strong structural ordering across graph
families and at graph sizes substantially beyond the training range, with distinct architecture-specific degradation profiles.
Adjacency-list serialization consistently outperforms edge-list encoding across all conditions, with the advantage widening at larger graph sizes. Per-property analysis reveals a
locality gradient in structural understanding, with node-level and
local-structural properties estimated reliably while global
combinatorial properties such as chromatic number represent a genuine
boundary of inference. Domain-specific fine-tuning yields strong in-distribution performance across biological, social network, chemical, and molecular graph domains.

Together, these findings map where fine-tuned small language models can be trusted for approximate graph reasoning and where they cannot, providing empirical grounding for their use as approximate structural reasoners in text-based workflows.




\section*{LLM Usage Statement}

This manuscript acknowledges the use of Claude Opus 4.6~\cite{claude} language model developed by Anthropic, to improve language clarity, refine sentence structure, and enhance overall writing precision.

\end{document}